\documentclass[a4paper]{article}
\usepackage{amsmath,graphicx}
\usepackage{color}
\usepackage{tabularx}
\usepackage{makecell}
\usepackage{url}
\usepackage{pifont}
\usepackage{INTERSPEECH2020}

\newcommand{\cmark}{\ding{51}}%
\newcommand{\xmark}{\ding{55}}%

\title{Semantic Mask for Transformer based End-to-End Speech Recognition}
\name{
	\makecell{Chengyi Wang$^{\diamondsuit}$, Yu Wu$^{\diamondsuit}$, \\ 
		(Alphabetical Order) \textit{ Yujiao Du$^{\ddagger}$\sthanks{Works are done during internship at Microsoft},  Jinyu Li$^{\dagger}$, Shujie Liu$^{\diamondsuit}$, Liang Lu$^{\dagger}$, Shuo Ren$^{\diamondsuit}$,}   \\ \textit{  Guoli Ye$^{\dagger}$, Sheng Zhao$^{\dagger}$, Ming Zhou$^{\diamondsuit}$}}}
\address{$^{\diamondsuit}$ Microsoft Research Asia, Beijing\\
	$^{\dagger}$Microsoft Speech and Language Group \\
	$^{\ddagger}$Beijing University of Posts and Telecommunications} 

\email{
	\{v-chengw, Wu.Yu, jinyli,  shujliu, lial, v-shure, guoye, Sheng.Zhao, mingzhou\}@microsoft.com \\
	magicbubble95@gmail.com}

\begin{document}

\maketitle
\begin{abstract}Attention-based encoder-decoder model has achieved impressive results for both automatic speech recognition (ASR) and text-to-speech (TTS) tasks. This approach takes advantage of the memorization capacity of neural networks to learn the mapping from the input sequence to the output sequence from scratch, without the assumption of prior knowledge such as the alignments. However, this model is prone to overfitting, especially when the amount of training data is limited. Inspired by SpecAugment and BERT, in this paper, we propose a semantic mask based regularization for training such kind of end-to-end (E2E) model. The idea is to mask the input features corresponding to a particular output token, e.g., a word or a word-piece, in order to encourage the model to fill the token based on the contextual information. While this approach is applicable to the encoder-decoder framework with any type of neural network architecture, we study the transformer-based model for ASR in this work. We perform experiments on Librispeech 960h and TedLium2 data sets, and achieve the state-of-the-art performance on the test set in the scope of E2E models.
	
\end{abstract}

\section{Introduction}
\label{sec:intro}

End-to-end (E2E) acoustic models, particularly with the attention-based encoder-decoder framework~\cite{DBLP:journals/corr/BahdanauCB14}, have achieved a competitive recognition accuracy in a wide range of speech datasets~\cite{DBLP:conf/icassp/PanayotovCPK15}. This model directly learns the mapping from the input acoustic signals to the output transcriptions without decomposing the problems into several different modules such as lexicon modeling, acoustic modeling and language modeling as in the conventional hybrid architecture. While this kind of E2E approach significantly simplifies the speech recognition pipeline, the weakness is that it is difficult to tune the strength of each component. One particular problem from our observations is that the attention based E2E model tends to make grammatical errors, which indicates that the language modeling power of the model is weak, possibly due to the small amount of training data, or the mismatch between the training and evaluation data. However, due to the jointly model approach in the attention model, it is unclear how to improve the strength of the language modeling power, i.e., attributing more weights to the previous output tokens in the decoder, or to improve the strength of the acoustic modeling power, i.e., attributing more weights to the context vector from the encoder. 



While an external language model may be used to mitigate the weakness of the language modeling power of an attention-based E2E model, by either re-scoring the hypothesis or through shallow or deep fusion~\cite{DBLP:conf/slt/ToshniwalKCWSL18}, the improvements are usually limited, and it incurs additional computational cost. Inspired by SpecAgument~\cite{DBLP:specaug} and BERT~\cite{DBLP:conf/naacl/DevlinCLT19}, we propose a semantic mask approach to improve the strength of the language modeling power in the attention-based E2E model, which, at the same time, improves the generalization capacity of the model as well. Like SpecAugment, this approach masks out partial of the acoustic features during model training. However, instead of using a random mask as in SpecAugment, our approach masks out the whole patch of the features corresponding to an output token during training, e.g., a word or a word-piece. The motivation is to encourage the model to fill in the missing token (or correct the semantic error) based on the contextual information with less acoustic evidence, and consequently, the model may have a stronger language modeling power and is more robust to acoustic distortions.

In principle, our approach is applicable to the attention-based E2E framework with any type of neural network encoder. To constrain our research scope, we focus on the transformer architecture~\cite{DBLP:conf/nips/VaswaniSPUJGKP17}, which is originally proposed for neural machine translation. Recently, it has been shown that the transformer model can achieve competitive or even higher recognition accuracy compared with the recurrent neural network (RNN) based E2E model for speech recognition~\cite{DBLP:journals/espnet}. Compared with RNNs, the transformer based encoder can capture the long-term correlations with a computational complexity of $O(1)$, instead of using many steps of back-propagation through time (BPTT) as in RNNs. We evaluate our transformer model with semantic masking on Librispeech and TedLium datasets. We show that semantic masking can achieve significant word error rate reduction (WER) on top of SpecAugment, and we report the lowest WERs on the test sets of the Librispeech corpus with an E2E model. We release our code, training scripts, and pre-train models at \url{https://github.com/MarkWuNLP/SemanticMask}      



\section{Related Work}

As aforementioned, our approach is closely related to SpecAugment~\cite{DBLP:specaug}, which applies a random mask to the acoustic features to regularize an E2E model. However, our masking approach is more structured in the sense that we mask the acoustic signals corresponding to a particular output token. Besides the benefit in terms of model regularization, our approach also encourages the model to reconstruct the missing token based on the contextual information, which improves the power of the implicit language model in the decoder. The masking approach operates as the output token level is also similar to the approach used in BERT~\cite{DBLP:conf/naacl/DevlinCLT19}, but with the key difference that our approaches works in the acoustic space. 

In terms of the model structure, the transformer-based E2E model has been investigated for both attention-based framework \cite{DBLP:conf/icassp/DongXX18,DBLP:journals/espnet} as well as RNN-T based models~\cite{DBLP:journals/corr/abs-1211-3711, zhang2020transformer}. Our model structure generally follows~\cite{DBLP:journals/corr/abs-1904-11660}, with a minor difference that we used a deeper CNN before the self-attention blocks. We used a joint CTC/Attention loss to train our model following~\cite{DBLP:journals/espnet}.

\section{Semantic Masking}
\label{sec:format}

\subsection{Masking Strategy}

\begin{figure}
	\centering
	\includegraphics[width=0.4\textwidth]{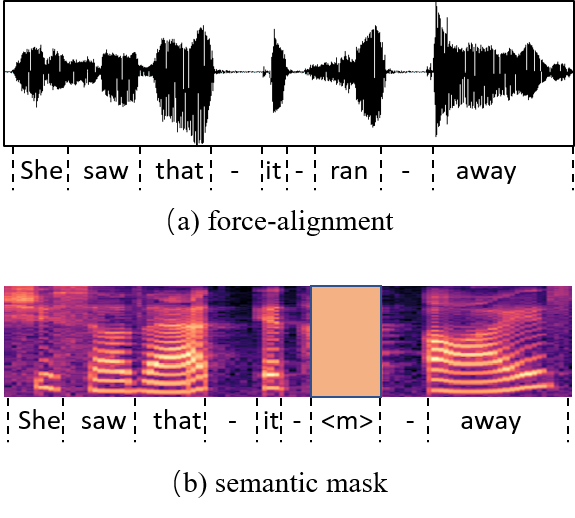}
	\caption{An example of semantic mask}
	\label{fig:semantic mask}
\end{figure}

Our masking approach requires the alignment information in order to perform the token-wise masking as shown in Figure \ref{fig:semantic mask}. There are multiple speech recognition toolkits available to generate such  alignments. In this work, we used the Montreal Forced Aligner\footnote{\url{https://github.com/MontrealCorpusTools/Montreal-Forced-Aligner}}
trained with the training data to perform forced-alignment between the acoustic signals and the transcriptions to obtain the word-level timing information. 
During model training, we randomly select a percentage of the tokens and mask the corresponding speech segments in each iteration. Following \cite{DBLP:conf/naacl/DevlinCLT19}, in our work, we randomly sample 15\% of the tokens and set the masked piece to the mean value of the whole utterance.

It should be noted that the semantic masking strategy is easy to combine with the previous SpecAugment masking strategy. Therefore, we adopt a time warp, frequency masking and time masking strategy in our masking strategy.  

\subsection{Why Semantic Mask Works?}
Spectrum augmentation \cite{DBLP:specaug} is similar to our method, since both  propose to mask spectrum for E2E model training. However, the intuitions behind these two methods are different. SpecAugment randomly masks spectrum in order to add noise to the source input, making the E2E ASR problem harder and prevents the over-fitting problem in a large E2E model. 

In contrast, our model aims to force the decoder to learn a better language model. Suppose that if a few words' speech features are masked, the E2E model has to predict the token based on other signals, such as tokens that have generated or other unmasked speech features. In this way, we might alleviate the over-fitting issue that generating words only considering its corresponding speech features while ignoring other useful features.  We believe our model is more effective when the input is noisy, because a model may generate correct tokens without considering previous generated tokens in a noise-free setting but it has to consider other signals when inputs are noisy, which is confirmed in our experiment.  

Moreover, our method slightly reduces the hyper-parameter tuning workload of SpecAugment and is more robust when  the variance of input audio length is large. 1) Time mask multiplicity and size of each mask are two hyper-parameters, which is required to be tuned based on different averaged utterance lengths and speaker speaking speeds. We believe that the hyper-parameter in our method, word masking ratio, is more stable with thre pre-computation alignment information.  2) A fixed masking strategy may be too large for short utterances, and too small for long utterances. Our method masks a utterance based on its total word count, addressing the length variance problem in traditional SpecAugment.  Concurrently, \cite{park2019specaugment}  proposes an adaptive SpecAug method to handle input audio with large length variance. 



\section{Model}
\label{sec:pagestyle}
Following \cite{DBLP:journals/corr/abs-1904-11660}, we add convolution layers before Transformer blocks and discard the widely used positional encoding component. According to our preliminary experiments, the convolution layers slightly improve the performance of the E2E model. In the following, we will describe the CNN layers and Transformer block respectively.

\subsection{CNN Layer}
\begin{figure}[t]		
	\begin{center}
		\includegraphics[width=.9\columnwidth]{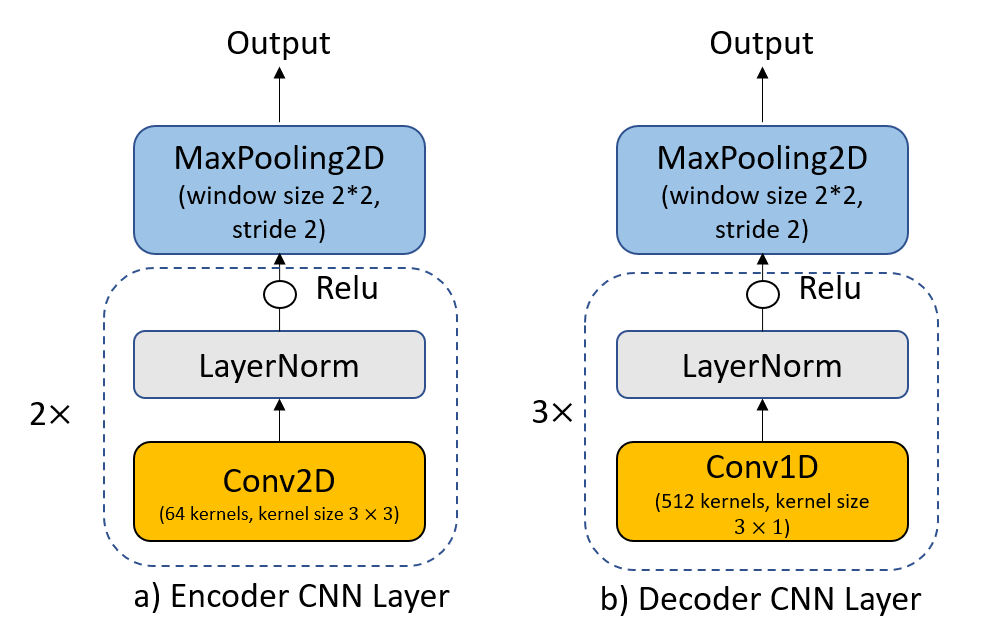}
	\end{center}
	
	\caption{CNN layer architecture.}\label{fig:cnn} 
	
\end{figure} 
We represent input signals as a sequence of log-Mel filter bank features, denoted as $\mathbf{X}=(x_0 \ldots, x_n)$, where $x_i$ is a  83-dim vector. 
Since the length of spectrum is much longer than text, we use VGG-like convolution block \cite{DBLP:journals/corr/SimonyanZ14a} with layer normalization and max-pooling function. The specific architecture is shown in Figure \ref{fig:cnn} . We hope the convolution block is able to learn local relationships within a small context and relative positional information. According to our experiments, the specific architecture outperforms Convolutional 2D subsampling method \cite{DBLP:journals/espnet}. We also use 1D-CNN in the decoder to extract local features replacing the position embedding \footnote{Experiment results show the Encoder CNN is more powerful than the decoder CNN.}.

\subsection{Transformer Block}
\begin{figure}[t]		
	\begin{center}
		\includegraphics[width=.9\columnwidth]{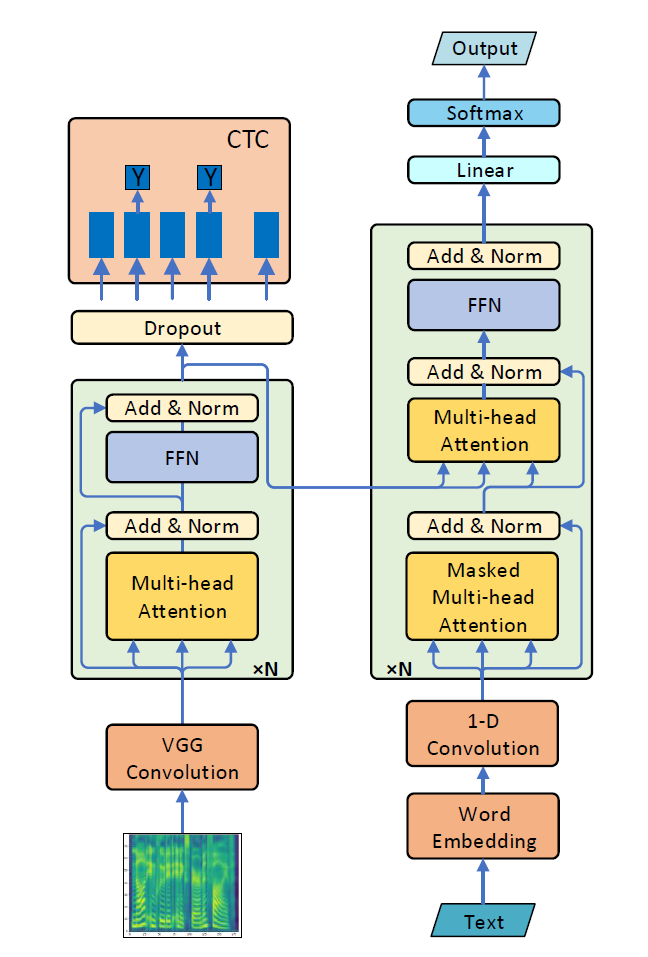}
	\end{center}
	
	\caption{E2E ASR model architecture.}\label{fig:transformer} 
	
\end{figure}

Our Transformer architecture is implemented as \cite{DBLP:journals/espnet}, depicting in Figure \ref{fig:transformer}. The transformer module consumes the outputs of CNN and extracts features with a self-attention mechanism. Suppose that $Q$, $K$ and $V$ are inputs of a transformer block, its outputs are calculated by the following equation
\begin{equation}
\text{SelfAttention}(\mathbf{Q,K,V})=\text{softmax}(\frac{\mathbf{QK}}{\sqrt{d_k}})\mathbf{V},
\end{equation} where $d_k$ is the dimension of the feature vector. To enable dealing with multiple attentions, multi-head attention is proposed, which is formulated as
\begin{flalign}
\text{Multihead}(\mathbf{Q,K,V}) &= [\mathbf{H_1} \ldots \mathbf{H}_{d_{head}}]\mathbf{W}^{head}\\
\text{where~}  \mathbf{H_i} &= SelfAttention(\mathbf{Q_i,K_i,V_i})
\nonumber
\end{flalign} where $d_{head}$ is the number of attention heads. Moreover, residual connection \cite{DBLP:conf/cvpr/HeZRS16}, feed-forward layer and layer normalization \cite{DBLP:journals/corr/BaKH16} are indispensable parts in Transformer, and their combinations are shown in Figure \ref{fig:transformer}.

\subsection{ASR Training and Decoding}
Following previous work \cite{DBLP:journals/espnet}, we employ a multi-task learning strategy to train the E2E model. Formally speaking, both the E2E model decoder and the CTC module predict the frame-wise distribution of $Y$ given corresponding source $X$, denoted as $P_{s2s}(\mathbf{Y}|\mathbf{X})$ and $P_{ctc}(\mathbf{Y}|\mathbf{X})$. We weighted averaged two negative log likelihoods to train our model
\begin{equation}
\mathcal{L} = - \alpha \log P_{s2s}(\mathbf{Y}|\mathbf{X}) - (1-\alpha) \log P_{ctc}(\mathbf{Y}|\mathbf{X}).
\end{equation} where $\alpha$ is set to 0.7 in our experiment. 

We combine scores of E2E model $P_{s2s}$, CTC score $P_{ctc}$ and a RNN based language model $P_{rnn}$ in the decoding process, which is formulated as
\begin{flalign}
\label{eq:pareto mle2}
P(y_i|\mathbf{X}, y_{<i})  =    &  P_{ctc}(y_i|\mathbf{X}, y_{<i})  + \beta_1 P_{s2s}(y_i|\mathbf{X}, y_{<i})\\
& + \beta_2 P_{rnn}(y_i|\mathbf{X}, y_{<i}),
\nonumber
\end{flalign} where $\beta_1$ and $\beta_2$  are tuned on the development set. 

We rescore our beam outputs based on another right-to-left language model $P_{r2l}(\mathbf{Y})$
and the sentence length penalty $\text{Wordcount}(\mathbf{Y})$.  The right-to-left language model is formulated as
\begin{equation}
P_{r2l}(\mathbf{Y}) = \prod_{i=n}^{1} P(y_i|y_{n}...y_{i+1}).
\end{equation}
It is a common practice to rerank outputs of a left-to-right s2s model with a right-to-left language model in the NLP community \cite{sennrich2017university}, since the right-to-left model is more sensitive to the errors existing in the right part of a sentence. We find its performance is better than a left-to-right Transformer language model consisting of more parameters. 

\begin{flalign}
\label{eq:pareto mle2}
Score(\mathbf{Y})  =      &  \log P_{s2s}(\mathbf{Y}|\mathbf{X})  + \gamma_1 \text{Wordcount}(\mathbf{Y})\\
& + \gamma_2 \log P_{r2l}(\mathbf{Y}). 
\nonumber
\end{flalign}
where $P_{trans\_lm}$ denotes the sentence generative probability given by a Transformer language model.

\section{EXPERIMENT}
\label{sec:page}
In this section, we describe our experiments on LibriSpeech \cite{DBLP:conf/icassp/PanayotovCPK15} and TedLium2 \cite{DBLP:conf/lrec/RousseauDE12}. We compare our results with state-of-the-art hybrid and E2E systems. We implemented our approach based on ESPnet \cite{DBLP:journals/espnet}, and the specific settings on two datasets are the same with \cite{DBLP:journals/espnet}, except the decoding setting. We use the beam size 20, $\beta_1 = 0.5$, and $\beta_2=0.7$ in our experiment. 

\begin{table}[t] \small
\scalebox{0.9}{	\begin{tabular}{l|l|l|l|l} 
		\hline
		&   \multicolumn{2}{c|}{Dev }    &        \multicolumn{2}{c}{Test }   \\ \hline
		& clean & other & clean & other \\ \hline
		\multicolumn{1}{l}{E2E Model}      \\ \hline
		RWTH (E2E)  \cite{luscher2019rwth}         & 2.9  &  8.4  &  2.8 & 9.3  \\ 
		LAS   \cite{DBLP:specaug}         & -  &  -  &  3.2  & 9.8  \\ 
		LAS+SpecAugment    \cite{DBLP:specaug}   & - & - & 2.5  &  5.8      \\ 
		ESPNET Transformer \cite{DBLP:journals/espnet}    & 2.2 & 5.6 & 2.6  &  5.7     \\ 
		Wav2letter Transformers \cite{synnaeve2019end}    &  2.56& 6.65& 3.05& 7.01    \\ 
		\,\,\,\, + LM Fusion \cite{synnaeve2019end}       & 2.11 & 5.25 & 2.30 & 5.64      \\ 
		\,\,\,\, + Rescore  \cite{synnaeve2019end}       & 2.17 & 4.67 & 2.31 & 5.18      \\ \hline
		Baseline Transformer & 3.51 & 9.10 & 3.69   & 8.95   \\
		\,\,\,\, + LM Fusion   &  2.40 & 6.02    &   2.66  &  6.15\\
		Base Model with SpecAugment & 3.33 & 9.05 & 3.57   &  9.00   \\
		\,\,\,\, + LM Fusion   & 2.20  & 5.73   &  2.39  & 5.94  \\
		Base Model with Semantic Mask  & 2.93  &  7.75  &  3.04  & 7.43  \\
		\,\,\,\, + LM Fusion  & 2.09  &  5.31 &  2.32  & 5.55  \\
		\,\,\,\,\,\,\,\,+ Speed  Perturbation   & 2.07  &  5.06  &  2.31  & 5.21 \\ \,\,\,\,\,\,\,\,\,\,\,\, +  Rescore  & 2.05 & 5.01  & 2.24  & 5.12 \\
		Large Model with Semantic Mask    & 2.64  &   6.90 &  2.74  & 6.65  \\
		\,\,\,\, + LM Fusion \& Speed   & 2.02  &  4.91 &  2.19  & 5.19  \\
		\,\,\,\,\,\,\,\,\,\,\,\, +  Rescore  & 1.98 &  4.78 &  2.08  & 4.95    \\ \hline
		\multicolumn{1}{l}{Hybrid Model }      \\ \hline
		RWTH (HMM)     \cite{luscher2019rwth}        & 1.9  &  4.5  &  2.3 & 5.0  \\ 
		Wang et al.   \cite{wang2019transformer}        & -  &  -  &  2.60 & 5.59  \\ 
		\,\,\,\, + Rescore       & -  &  -  &  2.26  & 4.85  \\ 
		Multi-stream self-attention \cite{DBLP:journals/multistream}       & 1.8 & 5.8 & 2.2 & 5.7    \\ \hline
	\end{tabular}}
	\caption{Comparison of the Librispeech ASR benchmark}
	\label{cascade}
\end{table}


\subsection{Librispeech 960h} 
We represent input signals as a sequence of  80-dim log-Mel filter bank with 3-dim pitch features \cite{DBLP:conf/icassp/GhahremaniBPRTK14}. SentencePiece is employed as the tokenizer, and the vocabulary size is 5000. We train a base model with 12 encoder layers, 6 decoders, and the attention vector size is 512 with 8 heads, containing 75M parameters. To explore the effect of a large model, we enlarge the model to 24 encoder layers and 12 decoder layer, obtaining a 138M parameter model. The hyper-parameters of SpecAugment follow \cite{DBLP:journals/espnet} for a fair comparison. We use Adam algorithm to update the model, and the warmup step is $25000$. The learning rate decreases proportionally to the inverse square root of the step number after the $25000$-th step. We train our model $40$ epochs (decrease epochs due to speed perturbation) on 4 P40 GPUs, which approximately costs 5 days to coverage. We also apply speed perturbation by changing the audio speed to 0.9, 1.0 and 1.1. Following \cite{DBLP:journals/espnet}, we average the last 5 checkpoints as the final model. Unlike \cite{luscher2019rwth} and  \cite{wang2019transformer}, we use the same checkpoint for test-clean and test-other dataset.

The RNN language model uses the released LSTM language model provided by ESPnet\footnote{\url{https://github.com/espnet/espnet/tree/master/egs/librispeech/asr1}}. The architecture of right-to-left language model is the same as the RNN model provided by ESPnet, but it models a sentence from a different direction. 


We evaluate our model in different settings. The baseline Transformer represents the model with position embedding. The comparison between baseline Transformer and our architecture (Model with SpecAugment) indicates the improvements attributed to the architecture. Model with semantic mask is we use the semantic mask strategy on top of  SpecAugment, which outperforms model with SpecAugment with a large margin in a no external language model fusion setting, demonstrating that our masking strategy helps the E2E model to learn a better language model. The gap becomes smaller when equipped with a language model fusion component, which further confirms our motivation in Section \ref{sec:intro}. Speed Perturbation does not help model performance on the clean dataset, but it is effective on  the test-other dataset. Rescore is beneficial to both test-clean and test-other datasets. Larger model  size (138M v.s. 75M) provides over 10\% performance gain when LM model is absent. 

As far as we know, our model is the best E2E ASR system on the Librispeech testset. Our model is built upon the code base of ESPnet, and achieves relative $10\%$ gains in a fair comparison (75M base model), due to the better architecture and masking strategy. The parameter number of the large model is smaller than Wav2letter Transformer but it shows better performance.  
 Comparing with hybrid methods, our model obtains a similar performance on the test-clean set, but is still worse than the best hybrid model on the test-other dataset. Furthermore, the improvement on test-other dataset is more significant. The reason might be that our semantic masking is more suitable on a noisy setting, because the input features are not reliable and the model has to predict the next token relying on previous ones and the whole context of the input.



We also analyze the performance of different masking strategies, showing in Table \ref{ablation}, where  all models are in model based setting and shallow fused with the RNN language model. According to the comparison between the second line and the third line, we find that the word masking is slightly better than the time masking if other SpecAugment operations are absent. When it combined with other SpecAugment methods, the word masking is still better than the time masking, indicating the superior of the semantic masking. We also attempt to mask phrases and add white-noise to masked regions, but there is no improvement according to several experiment comparisons. 
\begin{table}[]  \centering
\scalebox{0.9}{	\begin{tabular}{l|l|l|l|l|l|l|l} 
		\hline
		\multicolumn{4}{c|}{ } &   \multicolumn{2}{c|}{Dev }    &        \multicolumn{2}{c}{Test }   \\ \hline
		\makecell{Time \\ wrap} & \makecell{Time \\ mask} &\makecell{Freq \\ mask} & \makecell{Word \\ mask}   & clean & other & clean & other \\ \hline
		\xmark & \xmark & \xmark  & \xmark & 3.30 & 9.97 & 3.62  & 10.20      \\ 
		\xmark & \cmark & \xmark  & \xmark & 2.37& 6.94 & 2.63&7.02 \\
		\xmark & \xmark & \xmark  & \cmark& 2.38& 6.54 & 2.52&6.91 \\
		\cmark & \cmark & \cmark  & \xmark   & 2.20  & 5.73   &  2.39  & 5.94    \\ 
		\cmark & \xmark & \cmark  & \cmark     & 2.18  &  5.45  &  2.36  & 5.65 \\ 
		\cmark & \cmark & \cmark  & \cmark   & 2.09  &  5.31 &  2.32  & 5.55  \\ \hline
	\end{tabular}}
	\caption{Ablation test of different masking methods. The  fourth line is a default setting of SpecAugment. The fifth line uses word mask to replace random time mask, and the last line combines both methods on the time axis. \label{ablation}}
\end{table}

\begin{table}[h]
	\centering
	\begin{tabular}{l|l} 
		\hline
		&   Test   \\ \hline
		Kaldi (Chain + TDNN + Large LM) & 9.0  \\ \hline
		ESPnet RNN & 11.0  \\ 
		ESPnet Transformer & 10.4  \\ 
		\,\,\,\,+SpecAugment       & 8.9      \\ 
		\,\,\,\,\,\,\,\,+ LM  Fusion    & 8.1      \\ 
		\,\,\,\,+Semantic Mask & 8.5\\
		\,\,\,\,\,\,\,\,+ LM Fusion  & \bf{7.7} \\ \hline
	\end{tabular}
	\caption{Experiment results on TEDLIUM2.  \label{TED}}
	
\end{table}

\subsection{TedLium2}
To verify the generalization of the semantic mask, we further conduct experiments on TedLium2 \cite{DBLP:conf/lrec/RousseauDE14} dataset, which is extracted from TED talks. The corpus consists of 207 hours of speech data accompanying 90k transcripts. For a fair comparison, we use the same data-preprocessing method, Transformer architecture and hyperparameter settings as in \cite{DBLP:journals/espnet}. Our acoustic features are 80-dim log-Mel filter bank and 3-dim pitch features, which is normalized by the mean and the standard deviation for training set. The utterances with more than 3000 frames or more than 400 characters are discarded. The vocabulary size is set to 1000. 

The experiment results are listed in Table \ref{TED}, showing a similar trend as  the results in Librispeech dataset. Semantic mask is  complementary to specagumentation, which enables better S2S language modeling training in an E2E model, resulting in a relative 4.5$\%$ gain. The experiment proves the effectiveness of semantic mask on a different and smaller dataset. 


\section{Conclusion}
This paper presents a semantic mask method for E2E speech recognition, which is able to train a model to better consider the whole audio context for the disambiguation. Moreover, we elaborate a new architecture for E2E model, achieving state-of-the-art performance on the Librispeech test set in the scope of E2E models.


\bibliographystyle{IEEEtran}

\bibliography{refs}


\end{document}